%% file: ms.tex
\documentclass[runningheads]{llncs}
\usepackage{graphicx}
\usepackage{amsmath,amssymb}
\usepackage{color}

\newcommand*{\LargerCdot}{\raisebox{-0.75ex}{\scalebox{2.5}{$\cdot$}}}
\DeclareMathOperator*{\argmax}{argmax}

\usepackage{tabularx}
\newcolumntype{C}[1]{>{\centering\arraybackslash}p{#1}}
\usepackage{hhline}
\usepackage{multirow}

\usepackage{chapterbib}

\usepackage{algorithm}
\usepackage{algpseudocode}
\algrenewcommand\algorithmicrequire{\textbf{Input:}}
\algrenewcommand\algorithmicensure{\textbf{Output:}}

\usepackage{chngcntr}
\usepackage{subfigure}

\begin{document}
\title{Improving Spatiotemporal Self-Supervision\\by Deep Reinforcement Learning} 

\titlerunning{Improving Spatiotemporal Self-Supervision\\by Deep Reinforcement Learning}
\author{Uta B\"uchler\thanks{Indicates equal contribution} \and
Biagio Brattoli$^\star$ \and
Bj\"orn Ommer}

\authorrunning{U. B\"uchler, B. Brattoli, B. Ommer}

\institute{Heidelberg University, HCI / IWR, Germany\\
\email{\{firstname.lastname\}@iwr.uni-heidelberg.de}}

\maketitle
\begin{abstract}
Self-supervised learning of convolutional neural networks can harness large amounts of cheap unlabeled data to train powerful feature representations. As surrogate task, we jointly address ordering of visual data in the spatial and temporal domain. The permutations of training samples, which are at the core of self-supervision by ordering, have so far been sampled randomly from a fixed preselected set. Based on deep reinforcement learning we propose a sampling policy that adapts to the state of the network, which is being trained. Therefore, new permutations are sampled according to their expected utility for updating the convolutional feature representation. Experimental evaluation on unsupervised and transfer learning tasks demonstrates competitive performance on standard benchmarks for image and video classification and nearest neighbor retrieval.
\end{abstract}

\section{Introduction}
Convolutional neural networks (CNNs) have demonstrated to learn powerful visual representations from large amounts of tediously labeled training data \cite{alexnet}. However, since visual data is cheap to acquire but costly to label, there has recently been great interest in learning compelling features from unlabeled data. Without any annotations, self-supervision based on surrogate tasks, for which the target value can be obtained automatically, is commonly pursued \cite{shufflealearn,opn,oddoneout,cliqueCNN2,jigsaw,contextpred,timo,counting,ltmotion,imagetext,colorization17,deeppermnet,RotNet,journalMiAr}. In colorization \cite{colorization17}, for instance, the color information is stripped from an image and serves as the target value, which has to be recovered. Various surrogate tasks have been proposed, including predicting a sequence of basic motions \cite{ltmotion}, counting parts within regions \cite{counting} or embedding images into text topic spaces \cite{imagetext}.

The key competence of visual understanding is to recognize structure in visual data. Thus, breaking the order of visual patterns and training a network to recover the structure provides a rich training signal. This general framework of permuting the input data and learning a feature representation, from which the inverse permutation (and thus the correct order) can be inferred, is a widely applicable strategy. It has been pursued on still images \cite{jigsaw,jigsaw++,contextpred,deeppermnet,multiTask} by employing spatial shuffling of images (especially permuting jigsaws) and in videos \cite{shufflealearn,opn,oddoneout,ourcvpr} by utilizing temporally shuffled sequences. Since spatial and temporal shuffling are both ordering tasks, which only differ in the ordering dimension, they should be addressed jointly. 

We observe that there has been unused potential in self-supervision based on ordering: Previous work \cite{opn,oddoneout,jigsaw,jigsaw++,ourcvpr} has \emph{randomly} selected the permutations used for training the CNN. However, can we not find permutations that are of higher utility for improving a CNN representation than the random set? For instance, given a $3\times 3$ jigsaw grid, shuffling two neighboring image patches, two patches in faraway corners, or shuffling all patches simultaneously will learn structure of different granularity. 
Thus diverse permutations will affect the CNN in a different way. Moreover the effect of the permutations on the CNN changes during training since the state of the network evolves.
During learning we can examine the previous errors the network has made when recovering order and then identify a set of best suited permutations. Therefore, wrapped around the standard back-propagation training of the CNN, we have a reinforcement learning algorithm that acts by proposing permutations for the CNN training. To learn the function for proposing permutations we simultaneously train a policy and self-supervised network by utilizing the improvement over time of the CNN network as a reward signal.

\section{Related Work}
We first present previous work on self-supervised learning using one task or a combination of surrogate approaches. Then we introduce curriculum learning procedures and discuss meta-learning for deep neural network.

\textbf{Self-Supervised Representation Learning:} In self-supervision, the feature representation is learned indirectly by solving a surrogate task. For that matter, visual data like images \cite{transinvar,contextpred,colorization17,deeppermnet,splitbrain,gvsd,inpainting,multiTask,RotNet} or videos \cite{shufflealearn,opn,oddoneout,ltmotion,objectsmove,wang2015,transinvar,ourcvpr} are utilized as source of information, but also text \cite{imagetext} or audio \cite{audioTorralba}. In contrast to the majority of recent self-supervised learning approaches, Doersch et al. \cite{multiTask} and Wang et al. \cite{transinvar} combine surrogate tasks to train a multi-task network. Doersch et al. choose 4 surrogate tasks and evaluate a naive and a mediated combination of those. Wang et al., besides a naive multi-task combination of these self-supervision tasks, use the learned features to build a graph of semantically similar objects, which is then used to train a triplet loss. Since they combine heterogeneous tasks, both methods use an additional technique on top of the self-supervised training to exploit the full potential of their approach. Our model combines two directly related ordering tasks, which are complementary without the need of additional adjustment approaches.

\textbf{Curriculum Learning:} In 2009 Bengio et al.\cite{bengio2009curriculum} proposed curriculum learning (CL) to enhance the learning process by gradually increasing the complexity of the task during training. CL has been utilized by different deep learning methods \cite{graves2017automated,sumer2017self,chang2017active} with the limitation that the complexity of samples and their scheduling during training typically has to be established a priori. Kumar et. al \cite{kumar2010self} define the sample complexity from the perspective of the classifier, but still manually define the scheduling. In contrast, our policy dynamically selects the permutations based on the current state of the network.

\textbf{Meta-Learning for Deep Neural Networks:} Recently, methods have proposed ways to improve upon the classical training of neural networks by, for example, automatizing the selection of hyper-parameters \cite{learning2learn,wogradient,fewshotopt,NAS,NDF,bier}. Andrychowicz et al. \cite{learning2learn} train a recurrent neural network acting as an optimizer which makes informative decisions based on the state of the network. Fan et al. \cite{NDF} propose a system to improve the final performance of the network using a reinforcement learning approach which schedules training samples during learning. Opitz et al. \cite{bier} use the gradient of the last layer for selecting uncorrelated samples to improve performance. Similar to \cite{learning2learn,NDF,bier} we propose a method which affects the training of a network to push towards better performances. In contrast to these supervised methods, where the image labels are fixed, our policy has substantial control on the training of the main network since it can directly alter the input data by proposing permutations.
\begin{figure}[t]
\includegraphics[width=\textwidth]{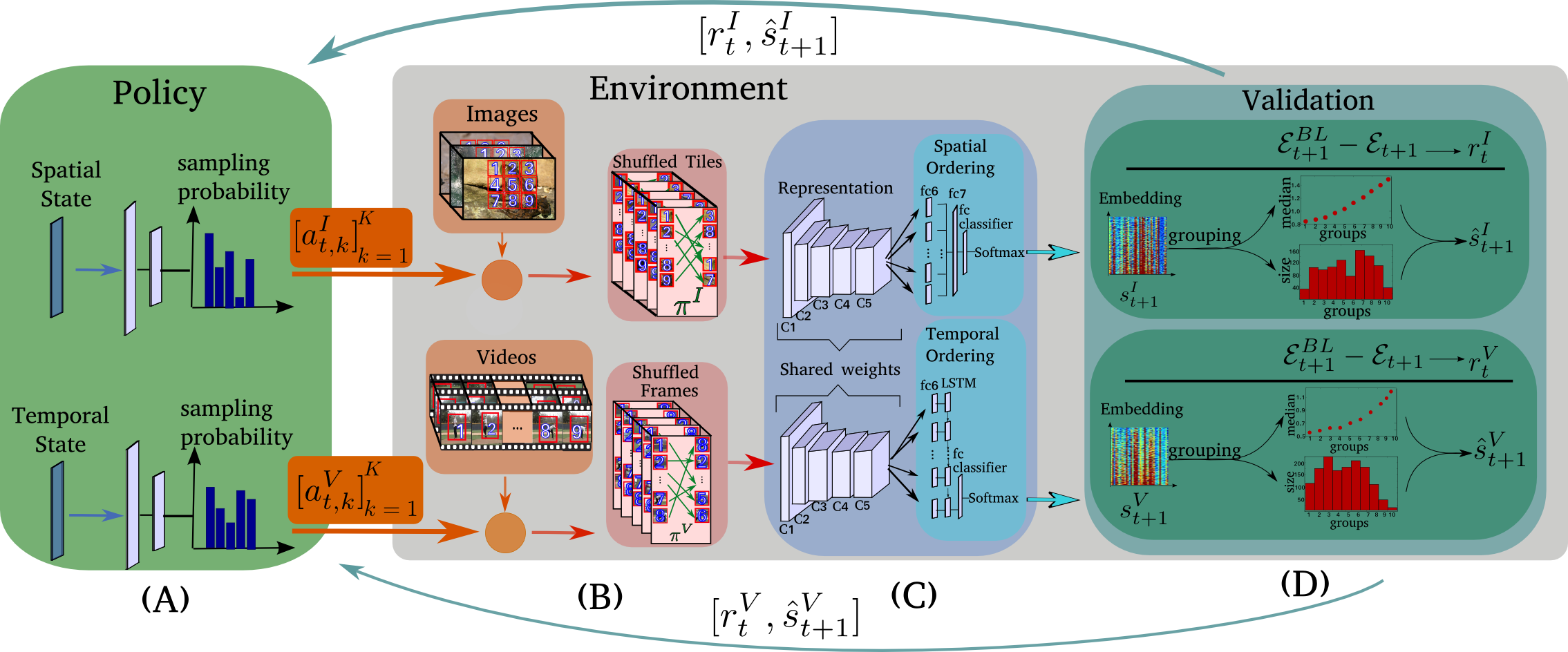}
\caption{(A) Deep RL of a policy for sampling permutations. (B) Permuting training images/videos by the proposed actions of (A) to provide self-supervision for our network architecture (C). (D) Evaluating the update network (C) on validation data to receive reward and state.}
\label{fig:ourmodel}
\end{figure}

\section{Approach}\label{sec:approach}
Now we present a method for training two self-supervised tasks simultaneously to learn a general and meaningful feature representation.  
We then present a deep reinforcement learning approach to learn a policy that proposes best suited permutations at a given stage during training.
\subsection{Self-Supervised Spatiotemporal Representation Learning}
Subsequently, we learn a CNN feature representation (CaffeNet \cite{caffenet} architecture up to pool5)
for images and individual frames of a video using spatiotemporal self-supervision (see Fig. \ref{fig:ourmodel}C). Training starts from scratch with a randomly initialized network. To obtain training samples for the spatial ordering task, we divide images into a $m \times m$ regular grid of tiles as suggested by \cite{jigsaw}(Fig. \ref{fig:ourmodel}B top). For temporal ordering of $u$ frames from a video sequence(Fig. \ref{fig:ourmodel}B bottom), shuffling is performed on frame level and with augmentation (detailed in Sect. \ref{sec:expSelfS}). Note that we do not require an object-of-interest detection, as for example done in \cite{shufflealearn,opn} by using motion (optical flow), since our approach randomly samples the frames from a video.

For the following part of this section, we are going to talk about a sample $x$ in general, referring to a sequence of frames (temporal task) or a partitioned image (spatial task). Let $x=\left(x_1,x_2,\dots \right)$ be the sample that is to be shuffled by permuting its parts by some index permutation $\psi_i = (\psi_{i,1},\psi_{i,2}, \cdots)$,
\begin{equation}
\psi_i(x) := \left(x_{\psi_{i,1}}, x_{\psi_{i,2}}, \dots \right) .
\end{equation}
The set of all possible permutations $\Psi^\star$ contains $u!$ or $(m\cdot m)!$ elements. If, for example, $u=8$ the total number of possible permutations equals $8!=40320$. For practical reasons, a pre-processing step reduces the set of all possible permutations, following \cite{jigsaw}, by sampling a set $\Psi \subset\Psi^\star$ of maximally diverse permutations $\psi_i \in \Psi$. We iteratively include the permutation with the maximum Hamming distance $d(\LargerCdot,\LargerCdot)$ to the already chosen ones.
Both self-supervised tasks have their own set of permutations.
For simplicity, we are going to explain our approach based on a general $\Psi$ without referring to a specific task.
To solve the ordering task of undoing the shuffling based on the pool5 features we want to learn (Fig. \ref{fig:ourmodel}(C)), we need a classifier that can identify the permutation. The classifier architecture begins with an fc6 layer. For spatial ordering, the fc6 output of all tiles is stacked in an fc7 layer; for temporal ordering the fc6 output of the frames is combined in a recurrent neural network implemented as LSTM \cite{lstm} (see Fig. \ref{fig:ourmodel}(C) and Sect. \ref{sec:expSelfS} for implementation details). The output of fc7 or the LSTM is then processed by a final fully connected classification layer. This last fc layer estimates the permutation $\psi_i$ applied to the input sample and is trained using cross-entropy loss. The output activation $ \varphi_i, i \in \{ 1, \dots |\Psi|\}$ of the classifier corresponds to the permutation $\psi_i \in \Psi$ and indicates how certain the network is that the permutation applied to the input $x$ is $\psi_i$. 
The network is trained in parallel with two batches, one of spatially permuted tiles and one of temporally shuffled frames. Back-propagation then provides two gradients, one from the spatial and one from the temporal task, which back-propagate through the entire network down to conv1.

The question is now, which permutation to apply to which training sample.
\subsection{Finding an Optimal Permutation Strategy by Reinforcement Learning}\label{sec:RL}
In previous works \cite{shufflealearn,opn,oddoneout,jigsaw,ourcvpr}, for each training sample one permutation is randomly selected from a large set of candidate permutations $\psi_i \in \Psi$. Selecting the data permutation independent from the input data is beneficial as it avoids overfitting to the training data (permutations triggered only by specific samples). However, permutations should be selected conditioned on the state of the network that is being trained to sample new permutations according to their utility for learning the CNN representation.

\textbf{A Markov Decision Process for Proposing Permutations:} We need to learn a function that proposes permutations conditioned on the network state and independent from samples $x$ to avoid overfitting. Knowingly, the state of the network cannot be represented directly by the network weights, as the dimensionality would be too high for learning to be feasible. To capture the network state at time step $t$ in a compact state vector $s$, we measure performance of the network on a set of validation samples $x \in X_{val}$.
Each $x$ is permuted by some $\psi_i \in \Psi$. A forward pass through the network then leads to activations $\varphi_i$ and a softmax activation of the network,
\begin{align} \label{eq:softmax}
y_i^\star &= \frac{exp(\varphi_{i})}{\sum_{k} exp(\varphi_{k})}.
\end{align}
Given all the samples, the output of the softmax function indicates how good a permutation $\psi_i$ can already be reconstructed and which ones are hard to recover (low $y_i^\star$). Thus, it reflects the complexity of a permutation from the view point of the network and $y_i^\star$ can be utilized to capture the network state $s$. To be precise, we measure the  network's confidence regarding its classification using the ratio of correct class $l$ vs. second highest prediction $p$ (or highest if the true label $l$ is not classified correctly):
\begin{align}\label{eq:softmaxRatio}
y_l(x) = \frac{y_l^\star(x)+1}{y_p^\star(x)+1},
\end{align}
where $x \in X_{val}$ and adding $1$ to have $0.5 \leq y_l \leq 2$, so that $y_l>1$ indicates a correct classification. The state $s$ is then defined as
\begin{align}\label{eq:state}
s = 
\begin{bmatrix}
    y_1(x_1) & \dots & y_1(x_{|X_{val}|}) \\
    \vdots & & \vdots\\
    y_{|\Psi|}(x_1) & \dots & y_{|\Psi|}(x_{|X_{val}|}),
\end{bmatrix}
\end{align}
where one row contains the softmax ratios of a permutation $\psi_i$ applied to all samples $x \in X_{val}$ (see Fig. \ref{fig:ourmodel}(D)). Using a validation set for determining the state has the advantage of obtaining the utility for all permutations $\psi_i$ and not only for the ones applied in the previous training phase. Moreover, it guarantees the comparability between validations applied at different time points independently by the policy.
The action $a=(x,\psi_i) \in A = X \times \Psi$ of training the network by applying a permutation $\psi_i$ to a random training sample $x$ changes the state $s$ (in practice we sample an entire mini-batch of tuples for one training iteration rather than only one). Training changes the network state $s$ at time point $t$ into $s'$ according to some transition probability $T(s^\prime|s,a)$.
To evaluate the chosen action $a$ we need a reward signal $r_t$ given the revised state $s^\prime$. The challenge is now to find \textit{the action} which maximizes the expected reward
\begin{align}\label{eq:MDPReward}
R(s,a) = \mathbb{E}[r_{t} | s_t = s,a],
\end{align}
given the present state of the network. The underlying problem of finding suitable permutations and training the network can be formulated as
a Markov Decision Process (MDP)\cite{sutton1998reinforcement}, a 5-tuple $<S,A,T,R,\gamma>$, where $S$ is a set of states $s_t$, $A$ is a set of actions $a_t$, $T(s^\prime|s,a)$ the transition probability, $R(a,s)$ the reward and $\gamma \in [0,1]$ is the discount which scales future rewards against present ones.

\textbf{Defining a Policy:} As a reward $r_t$ we need a score which measures the impact the chosen permutations have had on the overall performance in the previous training phase. For that, the error
\begin{align}
\mathcal{E} := 1 - \frac{1}{|\Psi|\cdot |X_{val}|}\sum\limits_{l=1}^{|\Psi|}\sum\limits_{x\in X_{val}} \delta_{\,l\,,\!\!\!\argmax\limits_{p=\{1,...,|\Psi|\}} \!\!y_p^\star(x)}
\end{align}
with $\delta$ the Kronecker delta, can be used to assess the influence of a permutation. 
To make the reward more informative, we compare this value against a baseline (BL), which results from simply extrapolating the error of previous iterations, i.e. $\mathcal{E}^{BL}_{t+1} = 2\mathcal{E}_t-\mathcal{E}_{t-1}$.
We then seek an action that improves upon this baseline.
Thus, the reward $r_t$ obtained at time point $t+1$ (we use the index $t$ for $r$ at time step $t+1$ to indicate the connection to $a_t$) is defined as
\begin{align} \label{eq:reward}
r_t := \mathcal{E}^{BL}_{t+1} - \mathcal{E}_{t+1}.
\end{align}
We determine the error using the same validation set as already employed for obtaining the state. In this way no additional computational effort is required.

Given the earlier defined state $s$ of the network and the actions $A$ we seek to learn a policy function
\begin{align}\label{eq:policy}
\pi(a|s,\theta) = P(a_t=a|s_t=s,\theta_t = \theta),
\end{align}
that, given the $\theta$ parameters of the policy, proposes an action $a=(x,\psi_i)$ for a randomly sampled training data point $x$ based on the state $s$, where $\pi(a|s,\theta)$ is the probability of applying action $a \in A$ at time point $t$ given the state $s$. The parameters $\theta$ can be learned by maximizing the reward signal $r$. It has been proven that a neural network is capable of learning a powerful approximation of $\pi$ \cite{DQN,sutton1998reinforcement,alphago}. However, the objective function (maximizing the reward) is not differentiable. In this case, Reinforcement Learning (RL)\cite{sutton1998reinforcement} has now become a standard approach for learning $\pi$ in this particular case.

\begin{figure}[t]
\centering
\includegraphics[width=0.80\textwidth]{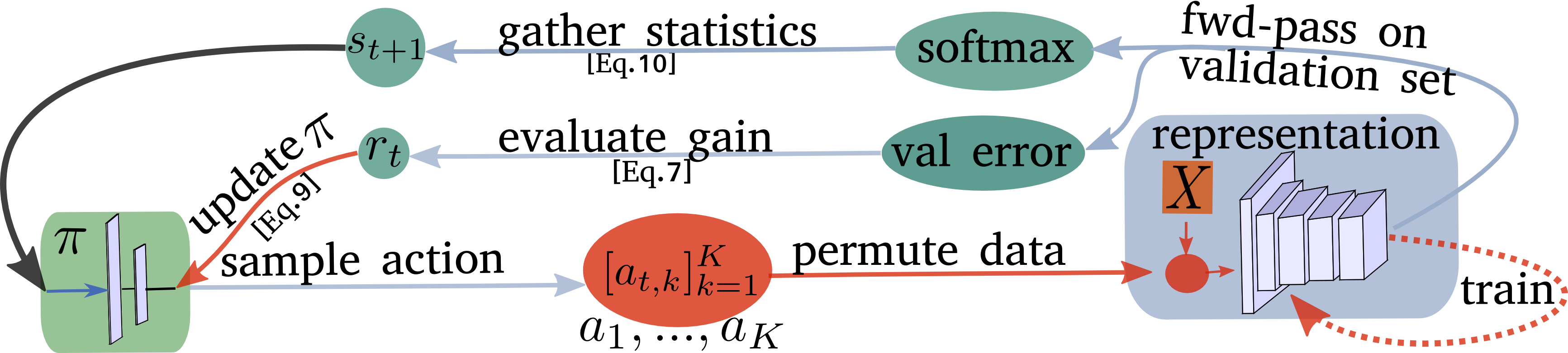}
\caption{Training procedure of $\pi$. The policy proposes actions $[a_{t,k}]^K_{k=1}$ to permute the data $X$, used for training the unsupervised network. The improvement of the network is then used as reward $r$ to update the policy.}
\label{fig:training}
\end{figure}
\textbf{Policy Gradient:} There are two main approaches for attacking deep RL problems: Q-Learning and Policy Gradient. We require a policy which models action probabilities to prevent the policy from converging to a small subset of permutations. Thus, we utilize a Policy Gradient (PG) algorithm which learns a stochastic policy and additionally guarantees convergence (at least to a local optimum) as opposed to Q-Learning. The objective of a PG algorithm is to maximize the expected cumulative reward (Eq. \ref{eq:MDPReward}) by iteratively updating the policy weights through back-propagation. One update at time point $t+1$ with learning rate $\alpha$ is given by
\begin{align}\label{eq:updatePolicy}
\theta_{t+1} = \theta_t + \alpha \Bigl( \sum\nolimits_{t^\prime \geq t}^{} \gamma^{t^\prime - t} r_{t^\prime}\Bigr) \nabla \log \pi(a|s,\theta),
\end{align}

\textbf{Action Space:} The complexity of deep RL increases significantly with the number of actions. Asking the policy to permute a sample $x$ given the full space $\Psi$ leads to a large action space. Thus, we dynamically group the permutations into $|C|$ groups based on the state of the spatiotemporal network. The permutations which are equally difficult or equally easy to classify are grouped at time point $t$ and this grouping changes over time according to the state of the network. 
We utilize the state $s$ (Eq. \ref{eq:state}) as input to the grouping approach, where one row $s_i$ represents the embedding of permutation $\psi_i$. 
A policy then proposes one group $c_j \in C$ of permutations and randomly selects one instance $\psi_i \in c_j$ of the group. Then a training data point $x$ is randomly sampled and shuffled by $\psi_i$. This constitutes an action $a=(x,\psi_i)$. Rather than directly proposing individual permutations $\psi_i$, this strategy only proposes a set of related permutations $c_j$. Since $|C|<<|\Psi|$, the effective dimensionality of actions is significantly reduced and learning a policy becomes feasible.

\textbf{Network State:} To obtain a more concise representation $\hat{s}=[\hat{s}_{j}]_{j=1}^{|C|}$ of the state of the spatiotemporal network (the input to the policy), we aggregate the characteristics of all permutations within a group $c_j$. Since the actions are directly linked to the groups, the features should contain the statistics of $c_j$ based on the state of the network. Therefore we utilize per group (i) the number of permutations belonging to $c_j$ and (ii) the median of the softmax ratios (Eq. \ref{eq:softmaxRatio}) over the $(\psi_i,x)$ pairs with $\psi_i \in c_j$ and $x \in X_{val}$
\begin{align}\label{eq:newState}
\hat{s} = [|c_j|,median\left([s_i]_{\psi_i\in c_j}\right)]_{j=1}^{|C|}.
\end{align}
The median over the softmax ratios reflects how well the spatiotemporal network can classify the set of permutations which are grouped together. Including the size $|c_j|$ of the groups helps the policy to avoid the selection of very small groups which could lead to overfitting of the self-supervised network on certain permutations. The proposed $\hat{s}$ have proven to be an effective and efficient representation of the state. Including global features, as for example the iteration or learning rate utilized in previous work \cite{learning2teach,NDF}, does not help in our scenario. It rather increases the complexity of the state and hinders policy learning. Fig. \ref{fig:ourmodel}(D) depicts the validation process, including the calculation of state $\hat{s}$ and the reward $r$.

\textbf{Training Algorithm:} We train the self-supervised network and the policy simultaneously, where the training can be divided in two phases: the self-supervised training and the policy update (see Fig. \ref{fig:training} and Algorithm 1 in section A of the Appendix). The total training runs for $T$ steps. Between two steps $t$ and $t+1$ solely the self-supervised network is trained ($\pi$ is fixed) using SGD for several iterations using the permutations proposed by $\pi$. Then, $\hat{s}$ is updated using the validation procedure explained above. At each time step $t$ an episode (one update of $\pi$) is performed.
During episode $t$, the policy proposes a batch of $K$ actions $[a_t]^K_{k=1}$, based on the updated state $\hat{s_t}$, which are utilized to train the self-supervised network for a small amount of iterations. At the end of the episode, another validation is applied to determine the reward $r_t$ for updating $\pi$ (Eq. \ref{eq:updatePolicy}). The two phases alternate each other until the end of the training.

\textbf{Computational Extra Costs during Training:} With respect to the basic self-super{\-}vised training, the extra cost for training the policy derives only from the total number of episodes $\times$ the time needed for performing an episode. If the number of SGD iterations between two policy updates $t$ and $t+1$ is significantly higher than the steps within an episode, the computational extra costs for training the policy is small in comparison to the basic training. Fortunately, sparse policy updates are, in our scenario, possible since the policy network improves significantly faster than the self-supervised network. 
We observed a computational extra cost of $\sim$40\% based on the optimal parameters. Previous work, \cite{learning2teach,NAS} which utilize deep RL for meta-learning, need to repeat the full training of the network several times to learn the policy, thus being several times slower.

\section{Experiments}\label{sec:exp}
In this section, we provide additional details regarding the self-supervised training of our approach which we evaluate quantitatively and qualitatively using nearest neighbor search. Then, we validate the transferability of our trained feature representation on a variety of contrasting vision tasks, including image classification, object detection, object segmentation and action recognition (Section \ref{sec:expSupervised}). We then perform an ablation study to analyze the gain of the proposed reinforcement learning policy and of combining both self-supervision tasks.

\subsection{Self-Supervised Training}\label{sec:expSelfS}
We first describe all implementation details, including the network architecture and the preprocessing of the training data. We then utilize two different datasets for the evaluation of the feature representation trained only with self-supervision.

\textbf{Implementation Details:}
Our shared basic model of the spatiotemporal network up to pool5 has the same architecture as CaffeNet \cite{caffenet} with batch normalization\cite{batchnorm} between the conv layers. To train the policy we use the Policy Gradient algorithm REINFORCE (with moving average subtraction for variance reduction) and add the entropy of the policy to the objective function which improves the exploration and therefore prevents overfitting (proposed by \cite{williams1991function}). The policy network contains 2 FC layers, where the hidden layer has 16 dimensions. We use K-means clustering for grouping the permutations in 10 groups. The validation set contains 100 ($|X_{val}|=100$) samples and is randomly sampled from the training set (and then excluded for training). The still images utilized for the spatial task are chosen from the training set of the Imagenet dataset \cite{imagenet}. For training our model with the temporal task, we utilize the frames from split1 of the human action dataset UCF-101 \cite{ucf}. We use 1000 initial permutations for both tasks ($|\Psi| = 1000$). Further technical details can be found in the Appendix, section B.

\begin{figure*}[t]
\centering
\includegraphics[width=\textwidth]{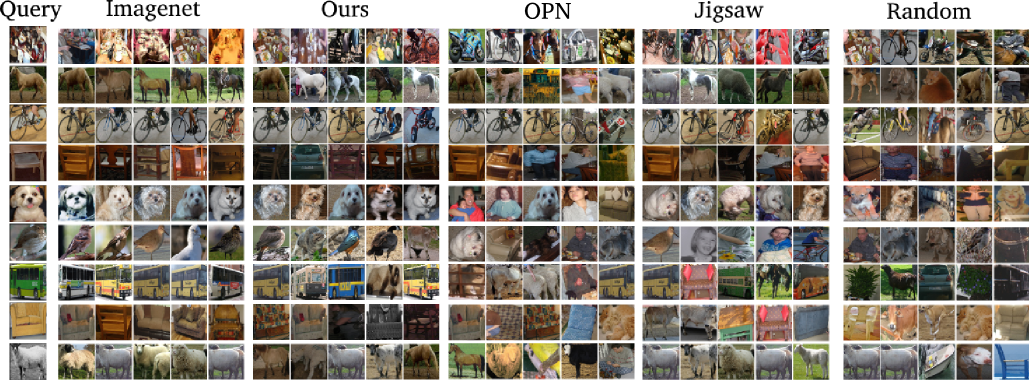}
\caption{Unsupervised evaluation of the feature representation by nearest neighbor search on the VOC07 dataset. For every test sample we show the Top5 nearest neighbors from the training set (Top1 to Top5 from left to right) using the cosine distance of the pool5 features. We compare the models from (i) supervised training with the Imagenet classification task, (ii) our spatiotemporal approach, (iii) OPN as a temporal approach \cite{opn}, (iv) Jigsaw as a spatial method \cite{jigsaw} and (v) a random initialization.}
\label{fig:NN}
\end{figure*}

\begin{table*}[t]
\begin{center}
\caption{Quantitative evaluation of our self-supervised trained feature representation using nearest neighbor search on split1 of UCF-101 and Pascal VOC 2007 dataset. Distance measure is cosine distance of pool5 features. For UCF101, 10 frames per video are extracted. Images of the test set are used as queries and the images of the training set as the retrieval targets. We report mean accuracies [\%] over all chosen test frames. If the class of a test sample appears within the top$k$ it is considered correctly predicted. We compare the results gained by (i) a random initialization, (ii) a spatial approach \cite{jigsaw}, (iii) a temporal method \cite{opn}, and (iv) our model. For extracting the features based on the weights of (ii) and (iii) we utilize their published models}\label{tab:unsuperEval}
\begin{tabular}{l|C{0.9cm}C{0.9cm}C{0.9cm}C{0.9cm}C{0.9cm}|C{0.9cm}C{0.9cm}C{0.9cm}C{0.9cm}C{0.9cm}C{0.9cm}}
\hline
\multirow{2}{*}{Methods}
&
\multicolumn{5}{c|}{UCF101}                                  
&                                            
\multicolumn{5}{c}{Pascal}
\\\cline{2-11} 
& Top1 & Top5 & Top10 & Top20 & Top50 & Top1 & Top5 & Top10 & Top20 & Top50 \\
\hhline{=|=====|=====}
Random & 18.8 & 25.7 & 30.0 & 35.0 & 43.3 & 17.6 & 61.6 & 75.5 & 85.5 & 94.2\\
\hline
Jigsaw \cite{jigsaw} & 19.7 & 28.5 & 33.5 & 40.0 & 49.4 & 39.2 & 71.6 & 82.2 & 89.5 & 96.0\\
OPN \cite{opn} & 19.9 & 28.7 & 34.0 & 40.6 & 51.6 & 33.2 & 67.1 & 78.5 & 87.0 & 94.6\\
Ours & \textbf{25.7} & \textbf{36.2} & \textbf{42.2} & \textbf{49.2} & \textbf{59.5} & \textbf{54.3} & \textbf{73.0} & \textbf{83.0} & \textbf{89.9} & \textbf{96.2}\\
\hline
\end{tabular}
\end{center}
\end{table*}
\textbf{Nearest Neighbor Search:}
To evaluate unsupervised representation learning, which has no labels provided, nearest neighbor search is the method of choice. For that, we utilize two different datasets: split1 of the human action dataset UCF-101 and the Pascal VOC 2007 dataset. UCF-101 contains 101 different action classes and over 13k clips. We extract 10 frames per video for computing the nearest neighbor.
The Pascal VOC 2007 dataset consists of 9,963 images, containing 24,640 annotated objects which are divided in 20 classes. Based on the default split, 50\% of the images belong to the training/validation set and 50\% to the testing set. We use the provided bounding boxes of the dataset to extract the individual objects, whereas patches with less than 10k pixels are discarded.
We use the model trained with our self-supervised approach to extract the pool5 features of the training and testing set and the images have an input size of 227$\times$227. Then, for every test sample we compute the Top$k$ nearest neighbors in the training set by using cosine distance. A test sample is considered as correctly predicted if its class can be found within the Top$k$ nearest neighbors. The final accuracy is then determined by computing the mean over all testing samples.
Tab. \ref{tab:unsuperEval} shows the accuracy for $k=1,5,10,20,50$ computed on UCF-101 and Pascal VOC 2007, respectively. It can be seen, that our model achieves the highest accuracy for all $k$, meaning that our method produces more informative features for object/video classification. Note, that especially the accuracy of Top1 is much higher in comparison to the other approaches.\\
We additionally evaluate our features qualitatively by depicting the Top5 nearest neighbors in the training set given a query image from the test set (see Fig. \ref{fig:NN}). We compare our results with \cite{jigsaw,opn}, a random initialization, and a network with supervised training using the Imagenet dataset.
\subsection{Transfer Capabilities of the Self-Supervised Representation}\label{sec:expSupervised}
Subsequently, we evaluate how well our self-trained representation can transfer to different tasks and also to other datasets. For the following experiments we initialize all networks with our trained model up to conv5 and fine-tune on the specific task using standard evaluation procedures.

\textbf{Imagenet \cite{imagenet}:} The Imagenet benchmark consists of $\sim$1.3M images divided in 1000 objects category. The unsupervised features are tested by training a classifier on top of the frozen conv layers. Two experiments are proposed, one introduced by \cite{colorization} using a linear classifier, and one using a two layer neural network proposed by \cite{jigsaw}. Tab. \ref{tab:imagenettest} shows that our features obtain more than 2\% over the best model with a comparable architecture, and almost 4\% in the linear task. The modified CaffeNet introduced by \cite{RotNet} is not directly comparable to our model since it has 60\% more parameters due to larger conv layers (groups parameter of the caffe framework\cite{caffenet}).

\begin{table}[!t]
\begin{minipage}[t]{.48\textwidth}
\begin{center}
\caption{Test accuracy [\%] of the Imagenet classification task. A Linear\cite{colorization} and Non-linear\cite{jigsaw} classifier are trained over the frozen features (pool5) of the methods shown in the left column.
(*: indicates our implementation of the model, +: indicates bigger architecture due to missing groups in the conv layers)}
\label{tab:imagenettest}
\begin{tabular}{l|C{1.5cm}C{1.5cm}}
\hline
Method & Non-Linear & Linear\\
\hhline{=|==}
Imagenet &  59.7 & 50.5\\
Random  & 12.0 & 14.1\\
\hline
RotNet+\cite{RotNet} & 43.8 & 36.5\\
\hline
Videos \cite{wang2015}  & 29.8 & -\\
OPN* \cite{opn}  & 29.6 & - \\
Context \cite{contextpred}  & 30.4 & 29.6\\
Colorization\cite{colorization} & 35.2 & 30.3\\
BiGan\cite{adversarial} & 34.8 & 28.0\\
Split-Brain\cite{splitbrain} & - & 32.8\\
NAT\cite{nat} & 36.0 & -\\
Jigsaw\cite{jigsaw} & 34.6 & 27.1\\
Ours & \textbf{38.2} & \textbf{36.5}\\
\hline
\end{tabular}
\end{center}
\end{minipage}
\quad
\begin{minipage}[t]{.48\textwidth}
\begin{center}
\caption{Transferability of features learned using self-supervision to action recognition. The network is initialized until conv5 with the approach shown in the left column and fine-tuned on UCF-101 and HMDB-51. Accuracies [\%] are reported for each approach. '*': Jigsaw (Noroozi et al. \cite{jigsaw}) do not provide results for this task, we replicate their results using our PyTorch implementation}
\label{tab:actionrec}
\begin{tabular}{l|C{1.5cm}C{1.5cm}}
\hline
Method & UCF-101 & HMDB-51 \\
\hhline{=|==}
Random & 47.8 & 16.3\\
Imagenet & 67.7 & 28.0 \\
\hline
Shuffle\&Learn \cite{shufflealearn} & 50.2 & 18.1 \\
VGAN \cite{gvsd} & 52.1 & -\\
Luo et. al \cite{ltmotion} & 53.0 & -\\
OPN \cite{opn} & 56.3 & 22.1\\
\hline
Jigsaw* \cite{jigsaw} & 51.5 & 22.5\\
Ours & \textbf{58.6} & \textbf{25.0}\\
\hline
\end{tabular}
\end{center}
\end{minipage}
\end{table}
\textbf{Action recognition:} For evaluating our unsupervised pre-trained network on the action recognition task we use the three splits of two different human action datasets: UCF-101 \cite{ucf} with 101 different action classes and over 13k clips and HMDB-51 \cite{hmdb} with 51 classes and around 7k clips. The supervised training is performed using single frames as input, whereas the network is trained and tested on every split separately. If not mentioned otherwise, all classification accuracies presented in this paragraph are computed by taking the mean over the three splits of the corresponding dataset. For training and testing we utilize the PyTorch implementation \footnote{\url{https://github.com/yjxiong/temporal-segment-networks}} provided by Wang et al. \cite{TSN2016ECCV} for augmenting the data and for the finetuning and evaluation step, but network architecture and hyperparameters are retained from our model. 
Table \ref{tab:actionrec} shows that we outperform the state-of-the-art by 2.3\% on UCF-101 and 2.9\% on HMDB-51. During our self-supervised training our network has never seen videos from the HMDB-51 dataset, showing that our model can transfer nicely to another dataset.

\begin{table}[!t]
\begin{center}
\caption{Evaluating the transferability of representations learned using self-supervision to three tasks on Pascal VOC. We initialize the network until conv5 with the method shown in the left column and fine-tune for (i) multi-label image classification\cite{krahenbuhl}, (ii) object detection using Fast R-CNN \cite{fastrcnn} and (iii) image segmentation\cite{FCN}. (i) and (ii) are evaluated on PASCAL VOC'07, (iii) on PASCAL VOC'12. For (i) and (ii) we show the mean average precision (mAP), for (iii) the mean intersection over union (mIoU). The fine-tuning has been performed using the standard CaffeNet, without batch normalization and groups 2 for conv[2,4,5]. ('+': significantly larger conv layers)}
\label{tab:pascalSuperv}
\begin{tabular}{l|C{2.5cm}C{2.5cm}C{2.5cm}}
\hline
Method & Classification\cite{pascal} & Detection\cite{pascal} & Segmentation\cite{pascal12} \\
\hhline{=|===}
Imagenet & 78.2 & 56.8 & 48.0 \\
Random & 53.3 & 43.4 & 19.8 \\
\hline
RotNet\cite{RotNet}+ & 73.0 & 54.4 & 39.1 \\
\hline
OPN\cite{opn} & 63.8 & 46.9 & - \\
Color17\cite{colorization17} & 65.9 & - & 38.4 \\
Counting\cite{counting} & 67.7 & 51.4& 36.6\\
PermNet\cite{deeppermnet} &69.4& 49.5& 37.9\\
Jigsaw\cite{jigsaw} & 67.6 & \textbf{53.2} & 37.6 \\
Ours &\textbf{74.2} & 52.8 & \textbf{42.8} \\
\hline
\end{tabular}
\end{center}
\end{table}
\textbf{Pascal VOC:} We evaluate the transferability of the unsupervised features by fine-tuning on three different tasks: multi-class object classification and object detection on Pascal VOC 2007 \cite{pascal}, and object segmentation on Pascal VOC 2012 \cite{pascal12}. In order to be comparable to previous work, we fine-tuned the model without batch normalization, using the standard CaffeNet with groups in conv2, conv4 and conv5. Previous methods using deeper networks, such as \cite{transinvar,multiTask}, are omitted from Table \ref{tab:pascalSuperv}. For object classification we fine-tune our model on the dataset using the procedure described in \cite{krahenbuhl}. We do not require the pre-processing and initialization method described in \cite{krahenbuhl} for any of the shown experiments. For object detection we train Fast RCNN\cite{fastrcnn} following the experimental protocol described in \cite{fastrcnn}. We use FCN\cite{FCN} to fine-tune our features on the segmentation task. The results in Table \ref{tab:pascalSuperv} show that we significantly improve upon the other approaches. Our method outperforms even \cite{RotNet} in object classification and segmentation, which uses batch normalization also during fine-tuning and uses a larger network due to the group parameter in the conv layers.

\begin{figure}[t]
\begin{minipage}[t]{.48\textwidth}
  \includegraphics[width=\textwidth]{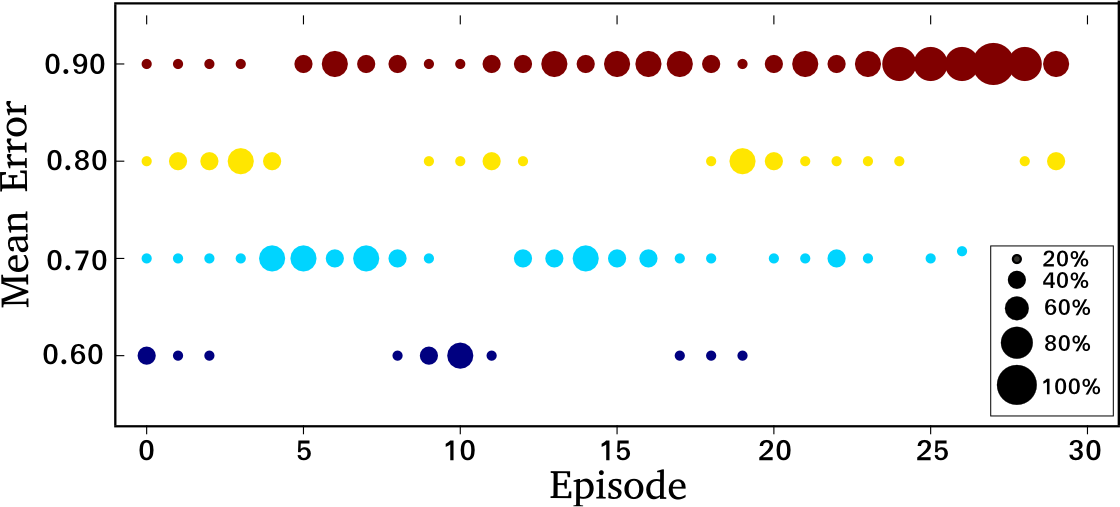}
  \caption{Permutations chosen by the policy in each training episode. For legibility, $\psi_i$ are grouped by validation error into four groups. The policy, updated after every episode, learns to sample hard permutations more often in later iterations}
  \label{fig:permutationsdensity}
\end{minipage}
\quad
\begin{minipage}[t]{.48\textwidth}
  \includegraphics[width=\textwidth]{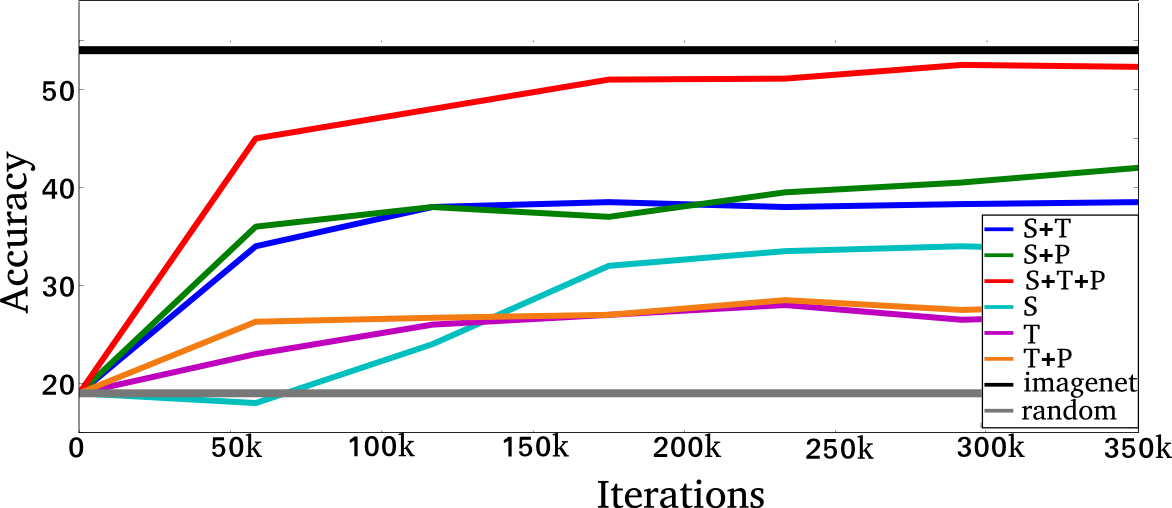}
  \caption{The test accuracy from Top1 nearest neighbor search evaluation on VOC07 is used for comparing different ablations of our architecture during training. The curves show a faster improvement of the features when the policy (P) is used}
  \label{fig:curveNN}
\end{minipage}
\end{figure}

\subsection{Ablation Study}
In this section, we compare the performances of the combined spatiotemporal (S+T) model with the single tasks (S,T) and show the improvements achieved by training the networks with the permutations proposed by the policy (P).

\begin{table}[!ht]
   \centering
\caption{We compare the different models on the multi-object classification task using the Pascal VOC07 and on the action recognition task using UCF-101. (S):\textbf{S}patial task, (T): \textbf{T}emporal task, (S+T):\textbf{S}patial and \textbf{T}emporal task simultaneously, (S+P):\textbf{S}patial task + \textbf{P}olicy, (S+T):\textbf{T}emporal task + \textbf{P}olicy, (S\&T):first solely \textbf{S}patial task, followed by solely \textbf{T}emporal task, (S+T+P):all approaches simultaneously}
\label{tab:ablation}
\begin{tabular}{l|C{1cm}C{1cm}C{1cm}C{1cm}C{1cm}C{1cm}C{1cm}C{1cm}}
\hline
Method & S & S+P & T & T+P & S\&T & S+T & S+T+P\\
\hhline{=|=======}
Pascal & 67.6 & 71.3 & 64.1 & 65.9 & 69.8 & 72.0 & \textbf{74.2}\\
UCF-101 & 51.5 & 54.6 & 52.8 & 55.7 & 54.2 & 57.3 & \textbf{58.6}\\
\hline
\end{tabular}
\end{table}

\textbf{Unsupervised Feature Evaluation:} In Fig. \ref{fig:curveNN} the models are evaluated on the Pascal VOC object classification task without any further fine-tuning by extracting pool5 features and computing cosine similarities for nearest neighbor search as described in section \ref{sec:expSelfS}. This unsupervised evaluation shows how well the unsupervised features can generalize to a primary task, such as object classification. Fig \ref{fig:curveNN} illustrates that the combined spatiotemporal model (S+T) clearly outperforms the networks trained on only one task (by 7\% on the spatial and 14\% on the temporal model). Furthermore, the combined network shows a faster improvement, which may be explained by the regularization effect that the temporal has on the spatial task and vice-versa. Fig. \ref{fig:curveNN} also shows, that each of the three models has a substantial gain when the CNN is trained using the policy. Our final model, composed of the spatiotemporal task with policy (S+T+P), reaches almost the supervised features threshold ("imagenet" line in Fig. \ref{fig:curveNN}).

\textbf{Supervised Fine-Tuning:} In Tab. \ref{tab:ablation}, a supervised evaluation has been performed starting from the self-supervised features. Each model is fine-tuned on the multi-class object classification task on Pascal VOC 2007 and on video classification using UCF-101. The results are consistent throughout the unsupervised evaluation, showing that the features of the spatiotemporal model (S+T) outperform both single-task models and the methods with RL policy (S+P and T+P) improve over the baseline models.
The combination of the two tasks has been performed in parallel (S+T) and in a serial manner (S\&T) by initializing the temporal task using the features trained on the spatial task. Training the permutation tasks in parallel provides a big gain over the serial version, showing that the two tasks benefit from each other and should be trained together.

\textbf{Policy Learning:} Fig. \ref{fig:permutationsdensity} shows the permutations chosen by the policy while it is trained at different episodes (x-axis). The aim of this experiment is to analyze the learning behavior of the policy. For this reason we initialize the policy network randomly and the CNN model from an intermediate checkpoint (average validation error 72.3\%). Per episode, the permutations are divided in four complexities (based on the validation error) and the relative count of permutations selected by the policy is shown per complexity. Initially the policy selects the permutations uniformly in the first three episodes, but then learns to sample with higher frequency from the hard permutations (with high error; top red) and less from the easy permutations (bottom purple), without overfitting to a specific complexity but mixing the hard classes with intermediate ones.

\begin{figure}[t]
\centering
\includegraphics[width=0.7\textwidth]{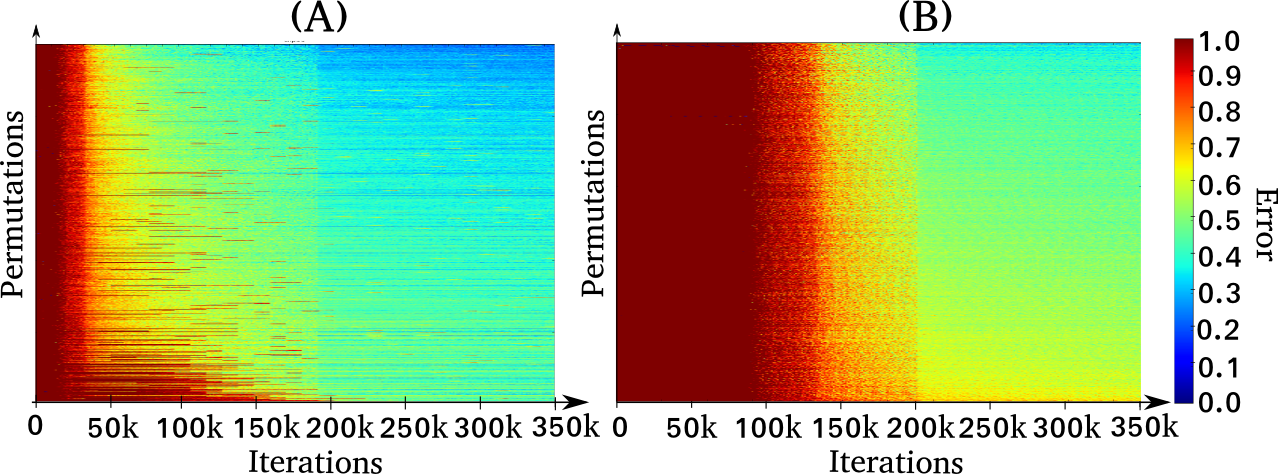}
\caption{Error over time of the spatial task, computed using the validation set and sorted by the average error. Each row shows how the error for one permutation evolves over time. (A): with Policy, (B): without policy}
\label{fig:error}
\end{figure}

Fig. \ref{fig:error} depicts the spatial validation error over the whole training process of the spatiotemporal network with and without the policy. The results are consistent with the unsupervised evaluation, showing a faster improvement when training with the permutations proposed by the policy than with random permutations. Note that (B) in Fig. \ref{fig:error} shows a uniform improvement over all permutations, whereas (A) demonstrates the selection process of the policy with a non-uniform decrease in error.

\section{Conclusion}
We have brought together the two directly related self-supervision tasks of spatial and temporal ordering. To sample data permutations, which are at the core of any surrogate ordering task, we have proposed a policy based on RL requiring relatively small computational extra cost during training in comparison to the basic training. Therefore, the sampling policy adapts to the state of the network that is being trained. As a result, permutations are sampled according to their expected utility for improving representation learning. In experiments on diverse tasks ranging from image classification and segmentation to action recognition in videos, our adaptive policy for spatiotemporal permutations has shown favorable results compared to the state-of-the-art.

{\let\thefootnote\relax\footnote{This work has been supported in part by DFG grant OM81/1-1, the Heidelberg Academy of Science, and an Nvidia hardware donation.}}
\clearpage

\bibliographystyle{splncs04}
\bibliography{egbib}

\include{appendix}
\end{document}

%% file: appendix.tex
%

\title{Appendix}

\titlerunning{Improving Spatiotemporal Self-Supervision by Deep Reinforcement Learning}
\authorrunning{Appendix}

\appendix
\counterwithin{figure}{section}
\counterwithin{table}{section}
\maketitle

\section{Algorithm for Updating the Policy}
Alg. \ref{alg:iteration} describes the training procedure for one episode $t$ in detail. Per episode, the weights $\theta$ of the policy network are updated using Eq.9 of the main paper. The state $\hat{s}$ based on the clustering $C$ and reward $r$ are computed using the function Validation($X_{\textrm{val}},\Phi$) following Section 3.2 (paragraph Network State) of the main paper. The function TrainCNN($\Phi,b$) performs one step (forward and backward pass) of the self-supervised network on the batch $b$ given the weights $\Phi$.\\
\begin{algorithm}[h!]
\caption{Episode $t$: policy network update}
\label{alg:iteration}
\begin{algorithmic}[1]
\Require $X$, $X_{\text{val}}$, $\pi$, $\theta$, $\Phi$, $\hat{s}$, $C$,$B$
\Comment{data, val set, policy, policy weights, CNN weights \\ \hspace{5.3cm} state, groups, batchsize}
\ForAll{$k = 1,...,K$}
\State {$b \leftarrow \emptyset$}
\Comment{set of permuted samples}
\State {$a \sim \pi_\theta(\hat{s})$}
\Comment{sample an action (a group)}
\Repeat
\State {$x \sim X$}
\Comment{random sample}
\State {$\psi \sim c_{a} \in C$}
\Comment{sample random permutation from chosen group}
\State {$b \leftarrow [b,\psi(x)$]}
\Comment{permute the sample}
\Until {$|b|==B$}
\State {$\Phi \leftarrow \textrm{TrainCNN}(\Phi,b)$} \Comment{Train the CNN given batch $b^\psi$}
\EndFor
\State {$\hat{s}, r, C \leftarrow $Validation($X_{\textrm{val}},\Phi$)} \Comment{Update $\hat{s}$,$C$ and compute $r$}
\State {$\theta \leftarrow$ Eq.9}
\Comment{Update policy weights using $r$}
\Ensure $\theta$, $s$, $\Phi$, $C$
\end{algorithmic}
\end{algorithm}

\section{Technical Details}
This section adds some further technical information to the implementation details mentioned in section 4.1 of the main paper. All deep networks are implemented using the PyTorch\footnote{\url{http://pytorch.org/}} framework. For the spatiotemporal network, we use SGD with a starting learning rate of 0.001 which we reduce after 200k iterations by a factor of 10. Our network runs in total for 350k iterations. We use a batchsize of 128 for both spatial and temporal tasks.
For the spatial classification branch, the fc6-layer has a size of 1024, fc7 has 4096 dimensions. For the temporal task we use an fc6-layer with 512 neurons and one LSTM layer with the hidden dimension of 256. The policy network is trained using the ADAM optimizer with a starting learning rate of 0.01.
For the temporal task, we randomly crop a patch with the size of 224x224 per frame and resize to 75x75. For the spatial task, each tile has the size of 75x75. Having the same input as the temporal task simplifies the implementation phase. As in \cite{jigsaw}, conv1 has stride 2 during the unsupervised training, and it is changed to 4 during all evaluation experiments. As augmentation, we randomly crop each tile/frame, apply a random color jittering to each of them and normalize the tiles/frames separately.
For the spatial task we divide an input image $x$ into 9 non-overlapping parts. For the temporal task, we randomly select 8 frames from each video. 

\section{Size of Validation Set}\label{sec:valsetsize}
\begin{figure}[t]
\centering
\includegraphics[width=0.5\textwidth]{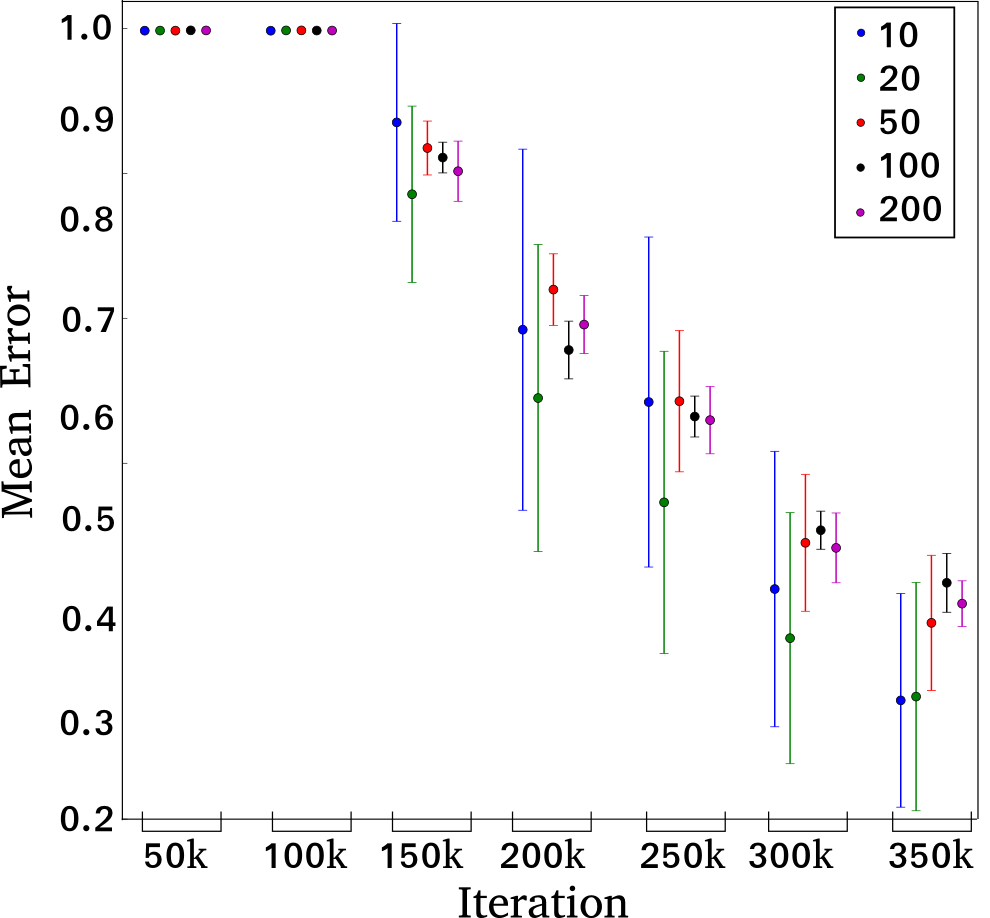}
\caption{Evaluation of the optimal size for $X_{val}$.}
\label{fig:sizeValSet}
\end{figure}
In Sect. 4.1 of the main paper, we mention the usage of 100 images for the validation set $X_{val}$. Fig. \ref{fig:sizeValSet} shows an evaluation of the optimal size for the validation set based on the mean and standard deviation of the error (y-axis) using several randomly sampled validation sets with size $|X_{val}| = 10,20,50,100$ or $200$ at different time steps (x-axis). We randomly sample 5 different sets per size and compute for every set the mean error given the checkpoints of the self-supervised network trained without policy at iteration 50k, 100k, 150k, 200k, 250k, 300k and 350k. Fig. \ref{fig:sizeValSet} then shows the mean and standard deviation over the 5 sets regarding a specific size and iteration. While the overall tendency of the error over the consecutive training iterations is similar for all validation sizes, $|X_{val}|=10,20,50$ show comparably large standard deviation. For the other two sizes there is only little difference which motivates our choice of $|X_{val}|=100$.

\section{Number of Groups}
\begin{figure}[t]%
\centering
\subfigure[Frequency of the softmax ratios $y_i^\star$ given all entries of $s$ (Eq. 2 of main submission) at time point $300k$.\label{fig:softmaxRatio}]{%
\includegraphics[width=0.45\textwidth]{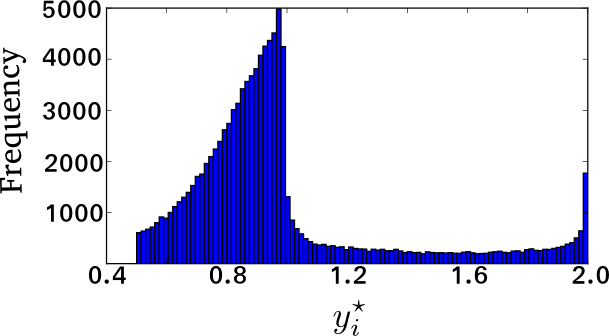}
}%
\qquad
\subfigure[Pairwise p-values for a different amount of groups at several checkpoints.\label{fig:nrGroups}]{%
\includegraphics[width=0.45\textwidth]{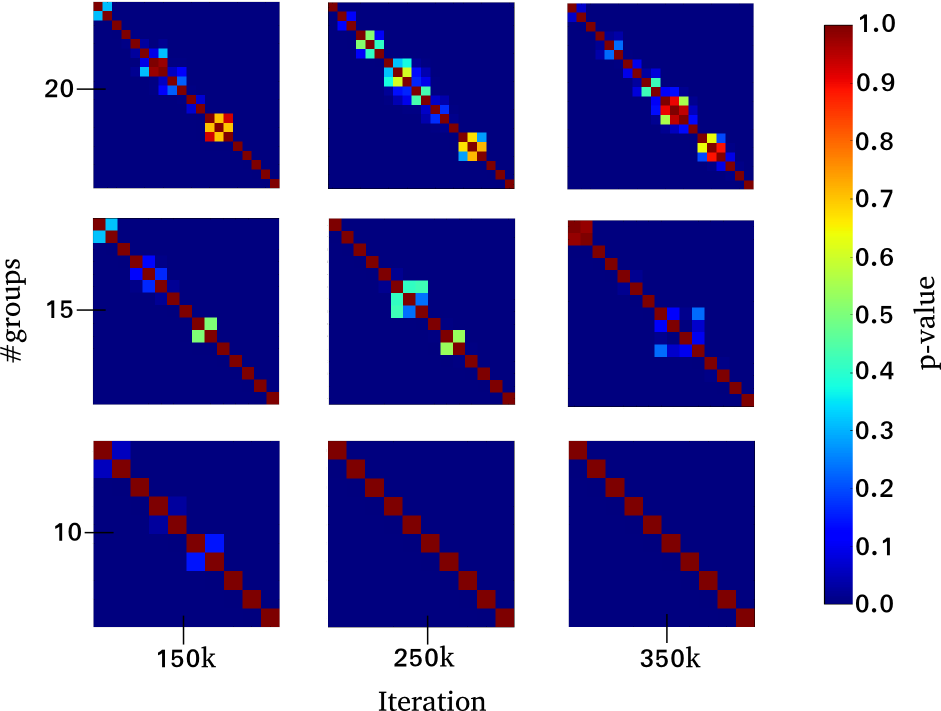}}%
\caption{Analysis of the optimal amount of groups.}
\end{figure}
We declare in Sect. 4.1 of the main paper the choice of 10 groups which we are going to analyze subsequently.
As defined in the main paper, we use the softmax ratios $y_i^\star$ (Eq. 3 of the main submission) to determine the complexity of a permutation from the view point of the network. Fig. B.\ref{fig:softmaxRatio} shows the distribution of all $y_i^\star$ over the $(\psi_i,x)$ pairs with $\psi_i \in \Psi$ and $x \in X_{val}$ (all entries of $s$ (Eq. 4 of main submission)) at time point $300k$ as histogram. We compute this distribution for all $\psi_i$ which are part of a group. We then test the distributions of the different groups for equality using the Kolmogorov-Smirnov-Test (KS-test; Null-Hypothesis is that the distributions are the same). If the p-value returned by the KS-test is smaller than a predefined significance value $\alpha=0.01$ the Null-Hypothesis can be rejected and the distributions are assumed to be different. We utilize this measure to identify groups which have a similar distribution and should therefore be grouped together. In this way, we can find the optimal amount of groups without having two separate groups with the same distribution/difficulty. Fig. B.\ref{fig:nrGroups} depicts the matrices of pairwise p-values for $|C|=10,15$ and $20$ at time point $150k$, $250k$ and $350k$. 
It can be seen, that there are already several groups for $|C|=15$ where the Null-Hypothesis cannot be rejected anymore (values higher than $\alpha$), i.e. $|C|=15$ is already to high for avoiding groups of similar complexity. Therefore, 10 groups seems to be the best choice for the clustering approach.

\section{Baseline Error $\mathcal{E}^{BL}$ Description}
\begin{figure}[h]
\centering
\includegraphics[width=0.50\textwidth]{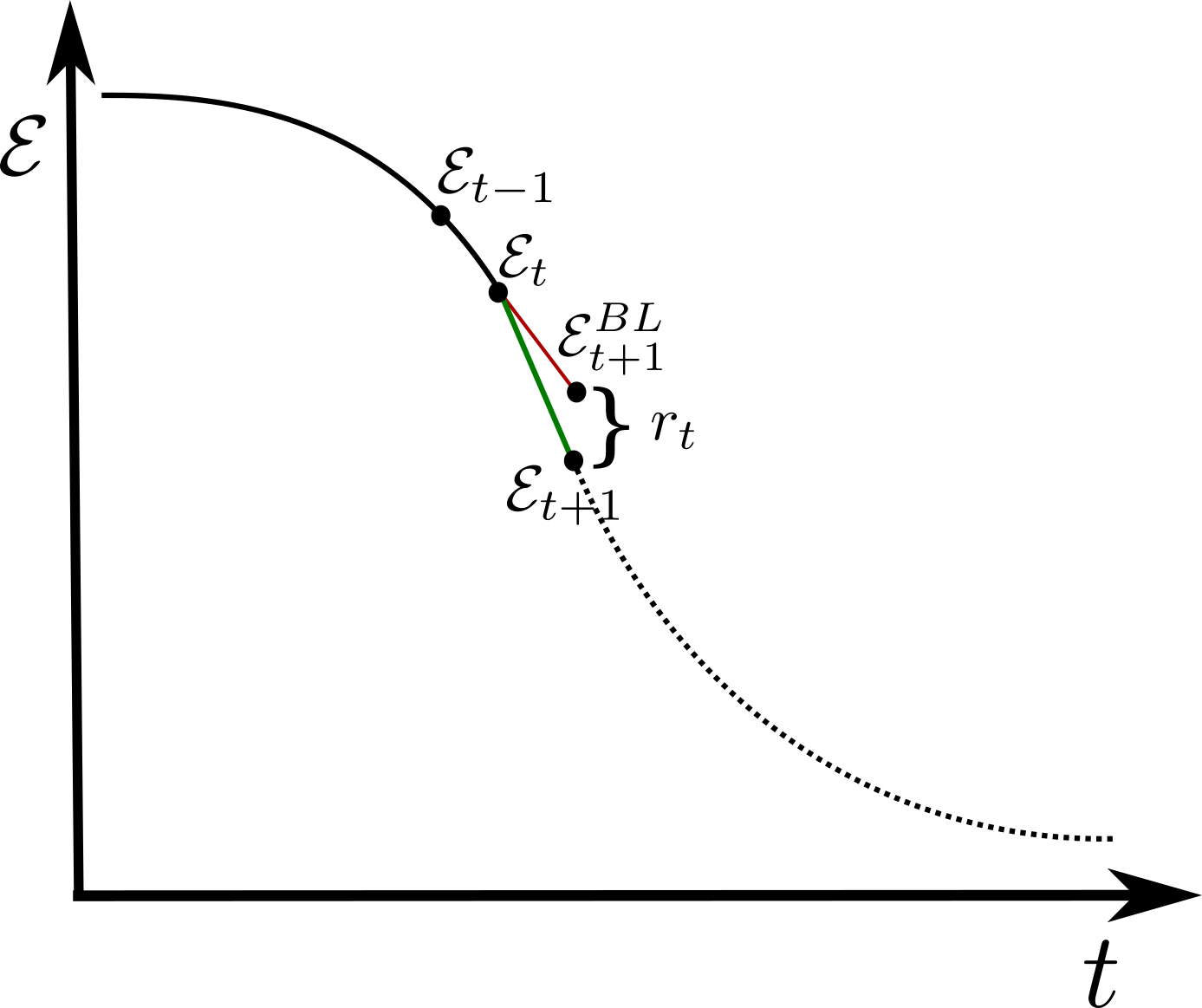}
\caption{The reward $r_t$ is positive when the error $\mathcal{E}_{t+1}$, obtained by training the self-supervised network using the policy, is below the extrapolated baseline error $\mathcal{E}^{BL}_{t+1}$}\label{fig:extrapolation}
\end{figure}

In Eq. 8 of the main paper we define the baseline error $\mathcal{E}^{BL}_{t+1}$ as the minimum error that the policy network needs to achieve to receive a positive reward. Fig.\ref{fig:extrapolation} illustrates more in details the use of this baseline with respect to the reward computation. The baseline $\mathcal{E}^{BL}_{t+1}$ is computed by linear extrapolation based on the error $\mathcal{E}_{t-1}$ and $\mathcal{E}_{t}$ in the previous time points $t-1$ and $t$. For extrapolating $\mathcal{E}^{BL}_{t+1}$ we use the equation
\begin{align}\label{eq:extrapolation}
f(u_3) = f(u_1) + \frac{u_3 - u_1}{u_2 - u_1} (f(u_2) - f(u_1)).
\end{align}
where a point $(u,f(u))$ corresponds to our errors $(t,\mathcal{E}_{t})$ and the extrapolated point $f(u_3)$ corresponds to our baseline error $\mathcal{E}^{BL}_{t+1}$. Substituting $\{u_1,u_2,u_3\}$ with $\{t-1,t,t+1\}$ and $f(u)$ with $\mathcal{E}_t$ results in
\begin{align}\label{eq:blerror}
\mathcal{E}^{BL}_{t+1} = \mathcal{E}_{t-1} + \frac{(t+1) - (t-1)}{t - (t-1)} (\mathcal{E}_{t} - \mathcal{E}_{t-1}) = 2\mathcal{E}_t-\mathcal{E}_{t-1}.
\end{align}

\clearpage
\section{How Decisive Are the Permutations?}
In this section we evaluate the impact on performance that permutation selection has during training. In particular, we use our trained policy for selecting the permutations and evaluate the model after one epoch. As baseline we use the random policy which selects the permutations uniformly at random. Moreover, we evaluate the permutations which are discarded by our policy. Therefore, we utilize an inverse policy in order to understand the importance of the permutation selection. It turns out that the inverse policy impairs training, producing features worse than the random policy (($78 \pm 16$)\%, see Table \ref{tab:selection}), while our policy always increases the performance with respect to the random policy.\\
The validation accuracy shown in Table \ref{tab:selection} refers to unsupervised training. We initialize the network from a given checkpoint and train for one epoch following one of the three policies. The final result is the ratio between our/inverse policy and the baseline random policy. Then we average over several checkpoints.

\begin{table}[t]
\begin{center}
\caption{Validation accuracy after one epoch of training using the policy in the left column. The accuracy is relative to the random policy. The experiment is repeated for several checkpoints, the reported accuracy is the mean and std over those repetitions. The performances when using the inverse policy are worse than the random policy baseline, while our policy always outperforms the baseline.}
\label{tab:selection}
\begin{tabular}{l|c}
\hline
Method & Relative Accuracy\\
\hline
RL policy& $(125\pm14)\%$\\
Inverse policy& $(78\pm16)\%$\\
\hline
\end{tabular}
\end{center}
\end{table}

\section{Permutation Selection of our Policy}
\begin{figure}[t]
\centering
\includegraphics[width=0.4\textwidth]{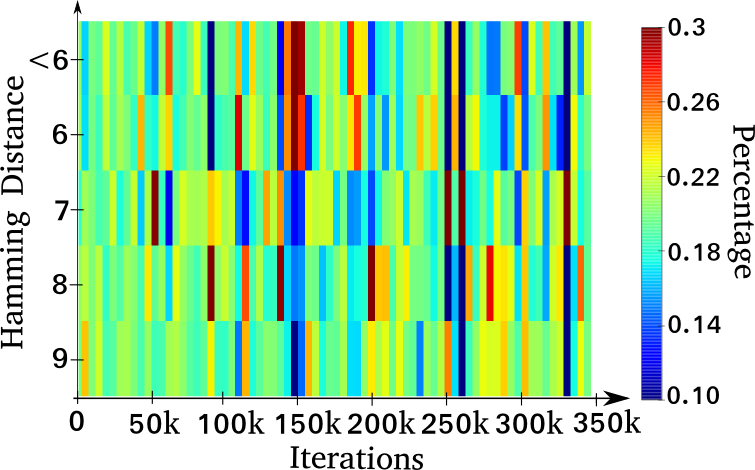}
\caption{Percentage of permutations chosen by the policy at diverse training iters. The permutations are structured using the Hamming distance to the not-shuffled sequence.}\label{fig:curriculumLearning}
\end{figure}

\begin{figure}[t]
\centering
\includegraphics[width=0.6\textwidth]{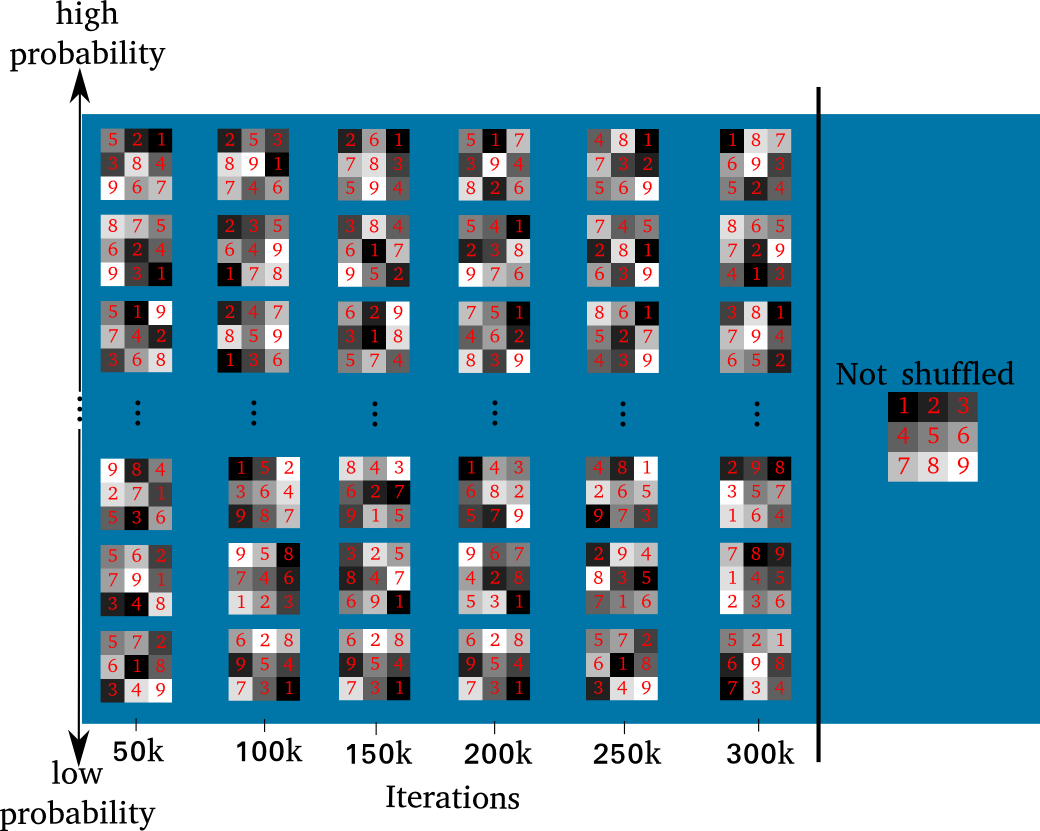}
\caption{Qualitative examples of permutations with a high or low probability to be chosen by the policy at different time points.}\label{fig:qualEx}
\end{figure}

As discussed in the introduction of the main paper, defining the complexity of a permutation should depend on the state of the network and not, for example, only on the degree of shuffling independently of the network. For this reason, we utilize the validation error as input for our policy. When illustrating how often permutations with a particular shuffling (Hamming distance to the not-shuffled sequence) are selected by our policy during the training process (see Fig. \ref{fig:curriculumLearning}) one should not be able to recognize a specific pattern, as for example a curriculum that selects easy samples (strongly permuted) at earlier iterations and harder ones (only small changes) at later iterations. Fig.  \ref{fig:curriculumLearning} shows that our trained policy does not follow a simple curriculum learning procedure. It selects the permutations only based on the state of the network as can be seen in Fig. 4 of the main article. Qualitative examples of chosen permutations, depicted in Fig. \ref{fig:qualEx}, confirm this behavior as no correlation between the degree of shuffling and the training iteration is visible.


\section{Extra Computational Costs}

\textbf{Policy Cost Calculation}\label{sec:cost}\\

\noindent In this section we derive the computational cost of using the policy during training relative to the computation of the basic self-supervised training. Including the policy introduces three additional phases in the training algorithm: action sampling (policy inference), update of the policy, and validation for computing state $\hat{s}$ and reward $r$. The inference and update of the policy are omitted from the calculation since their cost is orders of magnitude lower with respect to the main network, given the minor size of the policy network. Therefore the cost of the policy derives from the computational cost $V$ of the validation phase. In fact, for each sample in the validation set (100 samples following Sec.\ref{sec:valsetsize}), the main network performs one forward pass per each of the 1000 permutations, resulting in
\begin{align}\label{eq:v}
V = \frac{100 \cdot 1000}{128} \approx 780
\end{align}
where 128 is the batch size (Sec 4.1). The validation phase is then performed twice per episode $t$, at the beginning (computing $\hat{s}_t$) and the end of the episode (for the reward $r_t$). The final computational cost of using the policy is calculated by multiplying the episode cost by the number of episodes $T=90$ performed during the entire training
\begin{align}
CC = T \cdot (2\cdot V).
\end{align}
The number of total updates $T$ is set to $90$ since the policy does not benefit from an higher frequency of updates (no additional performance gain), which would only increase the computational cost.\\
We can compute the policy cost in relation to the self-supervised training as
\begin{align}
\frac{CC}{I} = \frac{T\cdot (2\cdot V)}{I} = \frac{90\cdot (2\cdot 780)}{350000} \approx 40\%.
\end{align}
given the total number of iterations $I = 350k$ to train the self-supervised network (Sec 4.1 in the main submission).\\
\begin{figure}[b]
\centering
\includegraphics[width=0.5\textwidth]{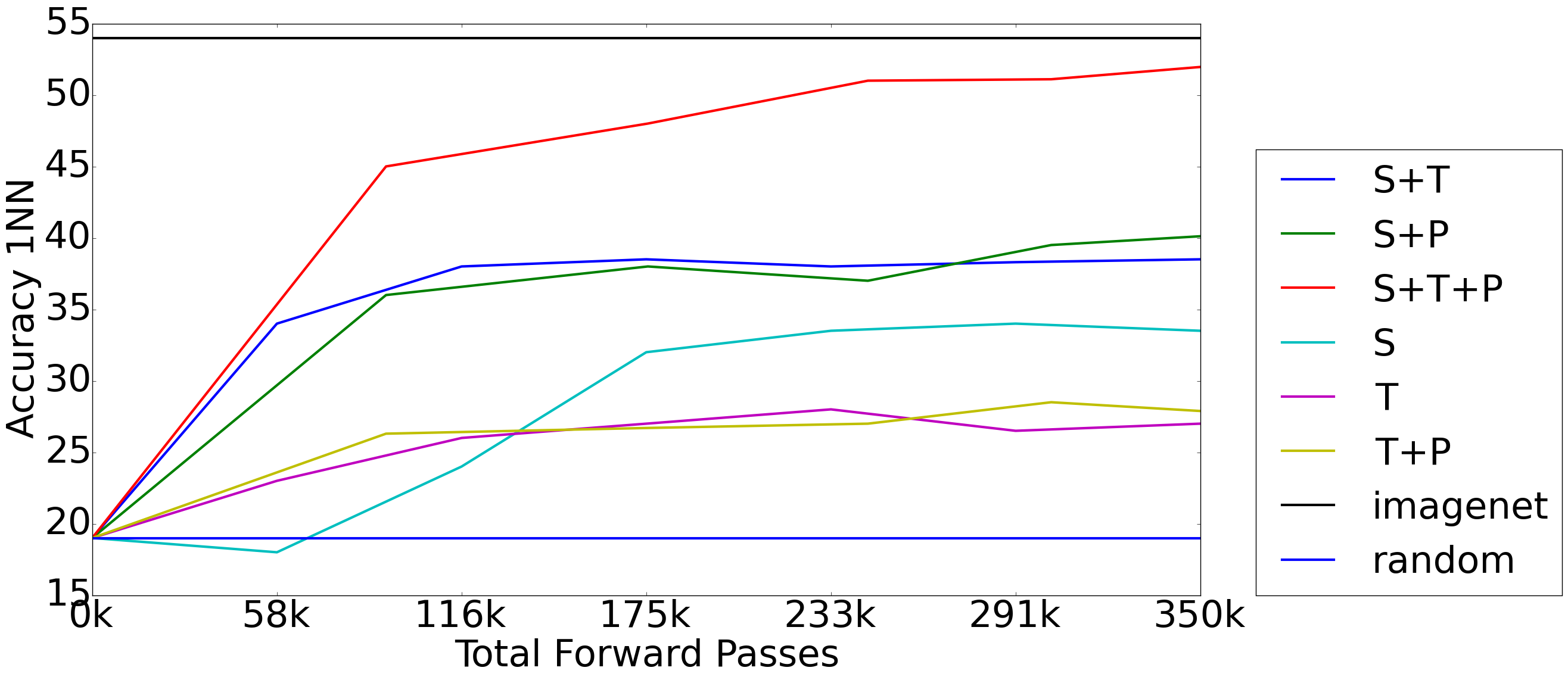}
\caption{Unsupervised object classification on Pascal VOC 2007 per number of total forward passes computed by the self-supervised network. The x-axes contains the unsupervised training plus the validation for the policy. 'Random' and 'Imagenet' are computed using respectively random weights and features trained with labels on ImageNet.}
\label{fig:computation}
\end{figure}

\noindent\textbf{Performance Relative to Computation}\\

\noindent Fig.5 in the main submission shows the performances of the unsupervised features during training, based on the iterations of the self-supervision network. Since our goal was to compare the convergence speed of the main network with and without policy, Fig.5 does not consider the additional iterations necessary for training the policy. For Fig. \ref{fig:computation} we normalize the x-axis by taking into account the episodes needed to train the policy network. Since the extra cost mainly derives from the forward passes of the self-supervised network during the validation phase, we use the total number of forward passes on the x-axis. Fig.\ref{fig:computation} shows that, even considering the extra computational cost needed to train the policy, there is a big advantage of using the policy during training.

\section{Visualizations}
\subsection{Activations}
\begin{figure}[h!]
\begin{minipage}[t]{.48\textwidth}
\centering
\includegraphics[width=0.7\textwidth]{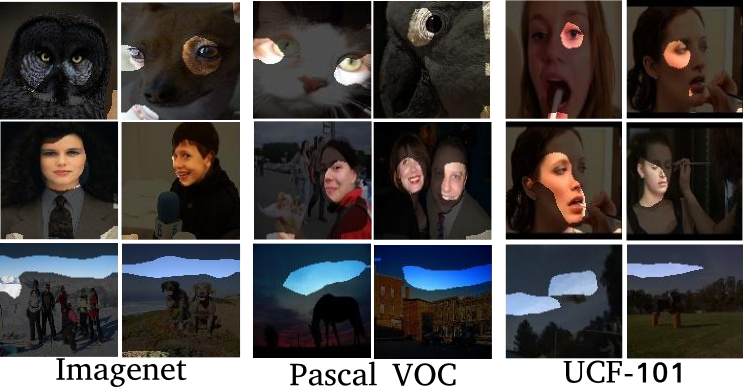}
\caption{Rows: Top activations of 3 different conv5 neurons across three datasets (columns). Note, that the neurons exhibit the same behavior in all datasets; The first unit focuses on eyes, the second on faces and the third on sky in a landscape.}
\label{fig:activation1}
\end{minipage}
\quad
\begin{minipage}[t]{.48\textwidth}
\centering
\includegraphics[width=0.7\textwidth]{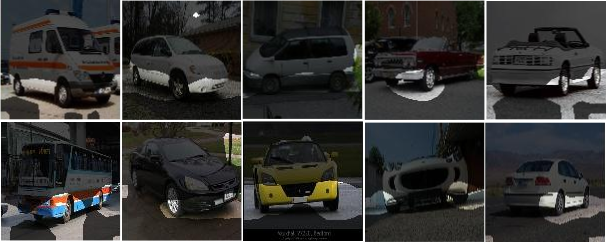}
\caption{Top activations of a single neuron firing on car wheels. The same neuron is evaluated on different images of Imagenet (1st row) and Pascal VOC (2nd row).}
\label{fig:activation2}
\end{minipage}
\end{figure}

Figure \ref{fig:activation1} and \ref{fig:activation2} show the top activations for different conv5 units of our self-supervised trained model following the approach described in \cite{activation}. In short, we run all images of a particular dataset through the network and output the top activations per unit contained in the conv5 layer. In Figure E.\ref{fig:activation1} we show three neurons over three different datasets: Imagenet, Pascal VOC, and UCF-101. Due to the number of images included in the Imagenet dataset, we only use the test set for visualizing the top activations. For UCF-101 we use one frame per video contained in the training set of split1. Having a consistent activation across different datasets shows the transfer capability of our feature representation. In particular the first row of Figure E.\ref{fig:activation1} is the activation of a unit responding to eyes, the second recognizes faces and the third reacts to sky in landscapes. Figure E.\ref{fig:activation2} shows additionally that the network learns to recognize very particular object parts, like the tires of a four-wheel vehicle.

\subsection{Characteristic Details}
\begin{figure}[h!]
\centering
\includegraphics[width=\textwidth]{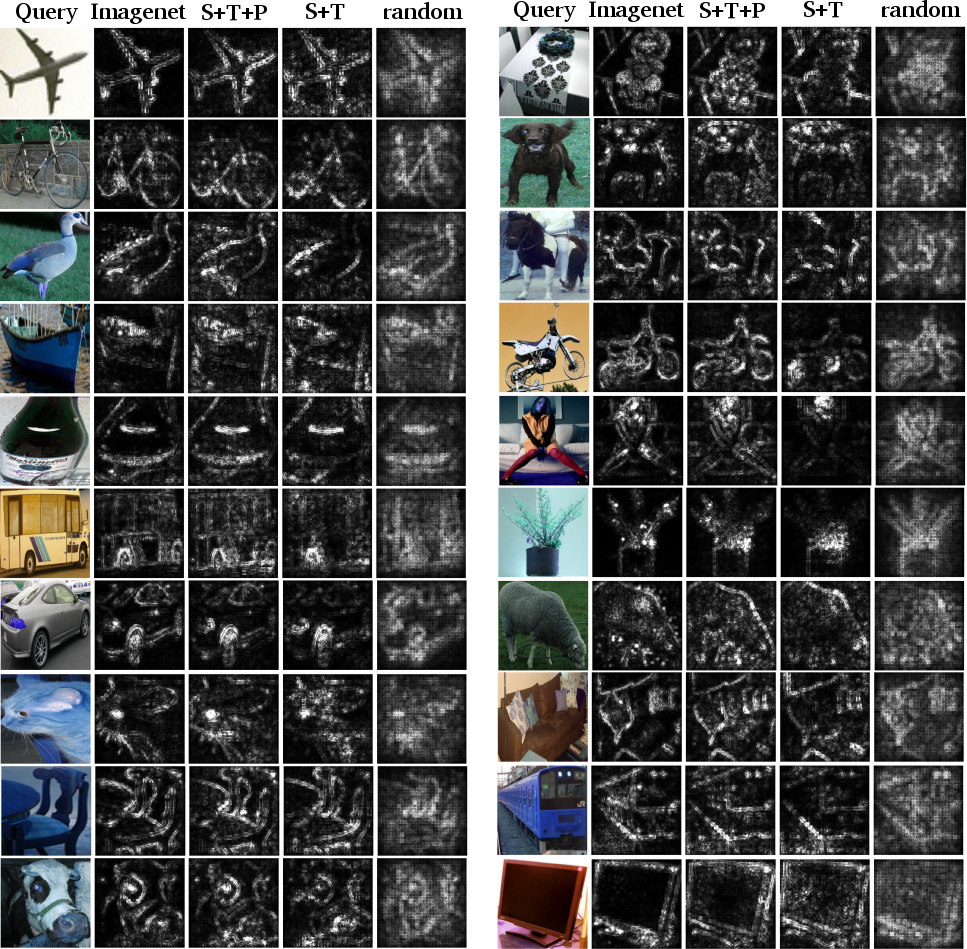}
\caption{Class Saliency Maps of image classification models using 4 different approaches as initialization.}
\label{fig:saliency}
\end{figure}

In Fig. \ref{fig:activation1} and \ref{fig:activation2} we have visualized individual neurons of the learned representation. Now we apply the visualization procedure of \cite{simonyan2013deep} to illustrate how well a representation has captured salient details of an object. As in Sect. 4.2 of the main submission, we evaluate our representation learned using self-supervision by transferring it to the task of image classification on Pascal VOC 2007. As described in Sect. 4.2, the network is initialized up to conv5 using (i) supervised training by means of the Imagenet classification task, (ii) our spatiotemporal self-supervision with the proposed sampling procedure (S+T+P, cf. Tab. 5), (iii) our approach without the sampling procedure (S+T), and (iv) no pre-training, i.e., random weights. Each of these initialized networks is then fine-tuned on Pascal VOC. Applying \cite{simonyan2013deep} yields a class-specific saliency map which indicates the relationship of individual pixels to the final classification. A good representation should, therefore, capture essential aspects of the object.  Fig. \ref{fig:saliency} shows that supervised pre-training on Imagenet using millions of images, which serves as an upper bound for our task, is followed by our self-supervised approach with sampling strategy (S+T+P). Without the policy, significantly more details are lost, as can be seen from (S+T). This underlines that the initial representation learned by our full model has captured more of the characteristic structure than the random permutations used in the literature. The last column highlights the importance of self-supervised pre-training compared to a random initialization.

\clearpage

\bibliographystyle{splncs04}
\bibliography{egbib_appendix}